%% file: main.tex
\documentclass[10pt,twocolumn,letterpaper]{article}

\usepackage{iccv}              


\definecolor{iccvblue}{rgb}{0.21,0.49,0.74}
\usepackage[pagebackref,breaklinks,colorlinks,allcolors=iccvblue]{hyperref}
\usepackage{multirow}
\usepackage{amssymb}
\usepackage{pifont}
\usepackage{bbding}
\usepackage{xcolor}

\def\paperID{} 
\def\confName{ICCV}
\def\confYear{2025}

\definecolor{yellow2}{RGB}{243, 230, 126}
\definecolor{green2}{RGB}{144,238,144}
\definecolor{red2}{RGB}{239, 177, 177}

\title{Warehouse Spatial Question Answering with LLM Agent\\
1st Place Solution of the 9th AI City Challenge Track 3
}

\author{
Hsiang-Wei Huang$^{1}$ \quad
Jen-Hao Cheng$^{1}$ \quad
Kuang-Ming Chen$^{1}$ \quad 
Cheng-Yen Yang$^{1}$ \quad \\
Bahaa Alattar$^{1}$ \quad
Yi-Ru Lin$^{1}$ \quad
Pyongkun Kim$^{2}$ \quad
Sangwon Kim$^{2}$ \quad
Kwangju Kim$^{2}$ \\
Chung-I Huang$^{3}$ \quad
Jenq-Neng Hwang$^{1}$ \\
\\
$^{1}$Information Processing Lab, University of Washington, USA\\
$^{2}$Electronics and Telecommunications Research Institute, South Korea\\
$^{3}$National Center for High-performance Computing, Taiwan \\
{\tt\small \{hwhuang, andyhci, kmchen, cycyang, balattar, yirulin, hwang\}@uw.edu},\\
{\tt\small \{iros, eddiekim, kwangju\}@etri.re.kr},\quad
{\tt\small 1203033@narlabs.org.tw} \\
\\
}

\begin{document}
\maketitle

\input{sec/0_abstract}    
\input{sec/1_intro}
\input{sec/2_survey}
\input{sec/3_method}
\input{sec/4_experiments}
\input{sec/6_conclusion}

\clearpage

{
    \small
    \bibliographystyle{ieeenat_fullname}
    \bibliography{main}
}

\end{document}

%% file: sec/0_abstract.tex
\begin{abstract}
Spatial understanding has been a challenging task for existing Multi-modal Large Language Models~(MLLMs). Previous methods leverage large-scale MLLM finetuning to enhance MLLM's spatial understanding ability. In this paper, we present a data-efficient approach. We propose a LLM agent system with strong and advanced spatial reasoning ability, which can be used to solve the challenging spatial question answering task in complex indoor warehouse scenarios. Our system integrates multiple tools that allow the LLM agent to conduct spatial reasoning and API tools interaction to answer the given complicated spatial question. Extensive evaluations on the 2025 AI City Challenge Physical AI Spatial Intelligence Warehouse dataset demonstrate that our system achieves high accuracy and efficiency in tasks such as object retrieval, counting, and distance estimation. The code is available at: \url{https://github.com/hsiangwei0903/SpatialAgent}.
\end{abstract}

%% file: sec/1_intro.tex
\input{figs/teaser}
\section{Introduction}
\label{sec:intro}
In recent years, the advancement of Large Language Models ~(LLM) has revolutionized LLM system development, especially on the 3D and spatial understanding fields~\cite{chen2024spatialvlm,cheng2025spatialrgpt,huang2025zero,zhang2024agent3d,zsvg3d,video3dllm,tosa}. A key component of these LLM agent systems is the ability to perceive, localize, and reason about various objects in the 3D scene. However, accurately estimating spatial relationships between objects and conducting complex spatial reasoning remain challenging for these systems, especially in complex indoor scenarios. Specifically, these systems focus mainly on addressing simpler tasks like view-selection~\cite{zhang2024agent3d} or perception and grounding tasks~\cite{xu2024vlmgrounder,li2025seeground}. The research of using an LLM agent to solve complex spatial reasoning problems using functions and tools remain underexplored by current methods.

On the other hand, some recent methods~\cite{chen2024spatialvlm,cheng2025spatialrgpt} directly conduct Multi-modal Large Language Model~(MLLM) training for spatial visual question answering tasks. These approaches rely on large-scale training data, which leads to extensive training cost. Moreover, the template-based training QA generation paradigm also limits their ability to conduct complicated spatial reasoning.

In this paper, we propose a spatial understanding agent system designed to robustly analyze object relationships in complex indoor warehouse scenarios. Our system leverages a reasoning LLM as an AI agent to conduct spatial reasoning, function calling, and question answering. We integrate a series of functions and models to interact with the agent. Several common tools and functions such as distance estimation and spatial relationship recognition—that together enable comprehensive spatial reasoning about object relationships and support high-level decision-making tasks.

We evaluate our system on the challenging 2025 AI City Challenge Physical AI Spatial Intelligence Warehouse benchmark. We show that our approach achieves state-of-the-art QA accuracy, providing a practical solution for warehouse spatial understanding systems and facilitate LLM agent research in the spatial understanding domain.

We summarize our contributions as follows:
\begin{itemize}
    \item We propose a spatial understanding LLM agent system with SoTA performance on warehouse spatial question answering tasks.
    \item Our designed LLM agent system possesses the ability to conduct complex spatial reasoning and further interact with multiple API tools, achieving data-efficient spatial question answering.
    \item We design multiple light-weight perception models and functions that can be used by the LLM agent, providing LLM agents active interaction and spatial reasoning over the given spatial query.
\end{itemize}

%% file: figs/teaser.tex
\begin{figure}[t]
    \centering
    \includegraphics[width=0.98\linewidth]{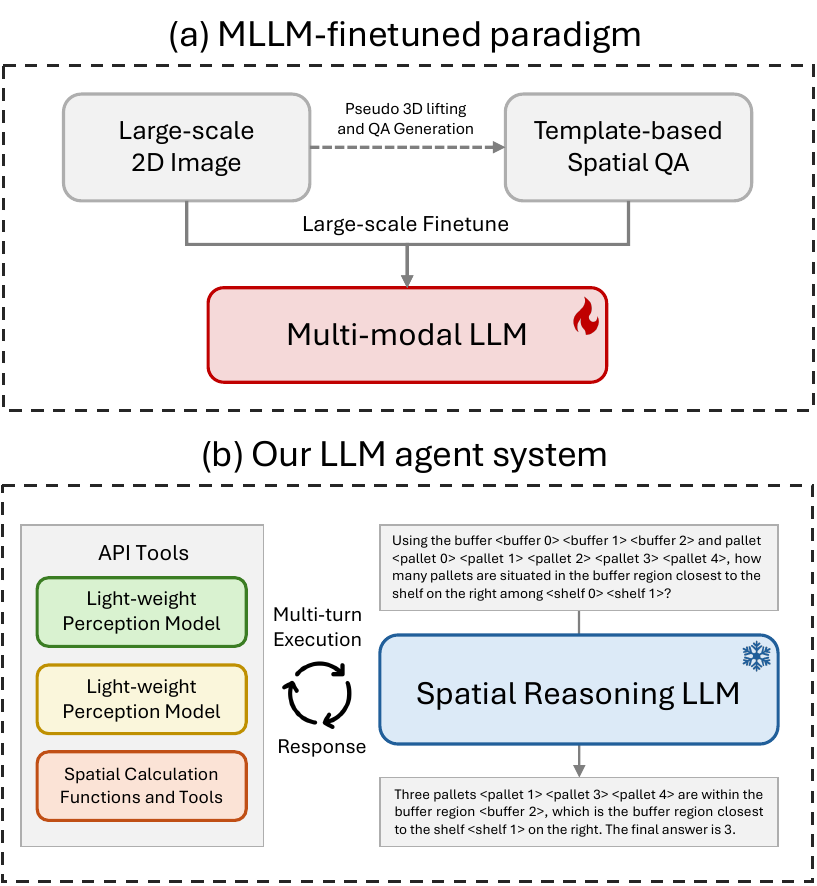}
    \caption{An illustration of (a) MLLM-finetuned paradigm like SpatialVLM~\cite{chen2024spatialvlm} or SpatialRGPT~\cite{cheng2025spatialrgpt}, which lift the 2D image to pseudo 3D point cloud and generate template-based QA pairs for large-scale MLLM finetuning. In contrast, (b) illustrates our proposed LLM agent system, which utilize a LLM agent that can conduct spatial reasoning and utilize multiple light-weight perception models and tools for complex spatial question answering task.}
    \label{fig:teaser}
\end{figure}

%% file: sec/2_survey.tex
\section{Related Work}
\label{sec:related_work}
\subsection{LLM Agent}
With the recent advancements of Large Language Models~(LLM), many research have explored using LLM as an agent. These LLM agent systems possess strong ability in language interaction, action planning, function calling, and interaction with tools. LLM agent system has demonstrated success for a wide range of tasks including video understanding~\cite{wang2024videoagent}, visual question answering~\cite{cai2024spatialbot}, embodied agent~\cite{zhao2024see,li2024embodied}, and 3D understanding~\cite{zhang2024agent3d}. Several recent works~\cite{zhang2024agent3d,xu2024vlmgrounder,llmgrounder} focus on solving the 3D and spatial question answering tasks using LLM agents, yet they mostly focus on more simple sub-tasks such as view selection~\cite{zhang2024agent3d,li2025seeground}, or only focus on visual grounding-related question~\cite{xu2024vlmgrounder,llmgrounder,zsvg3d}. The study of using LLM agents to perform both spatial reasoning and function calling to solve complex spatial understanding questions still remain underexplored. To this reason, we developed an intelligent LLM agent system that can perform both spatial reasoning and function calling to solve the challenging spatial question-answering task that involves diverse and complex natural language queries.

\subsection{Spatial Understanding MLLM}
Spatial understanding has been challenging for Multi-modal Large Language Models~(MLLM), as it requires inferring the 3D information from the 2D image and performing complex spatial reasoning. Existing works like SpatialVLM~\cite{cai2024spatialbot} and Spatial-RGPT~\cite{cheng2025spatialrgpt} leverage large-scale data for spatial understanding MLLM training, which incurs extensive collection and training cost. In this work, we leverage multiple light-weight perception models and a spatial reasoning-enabled LLM agent to achieve data-efficient spatial question answering.

%% file: sec/3_method.tex
\input{figs/agent}
\section{Method}
\label{sec:method}
\subsection{Spatial Agent}
Our spatial agent is designed to answer complex spatial questions by leveraging a Large Language Model~(LLM) with function-calling capabilities. The agent uses Gemini 2.5-Flash~\cite{comanici2025gemini}, and optionally supports its think mode configuration that activates Gemini's built-in reasoning budget to enhance multi-step reasoning performance.

Given an input image, a pair of binary masks, and a spatial question, the agent first parses and identifies relevant object masks using a rule-based parser. These masks are mapped to region identifiers and registered in the tool API. The question is then combined with a few-shot prompting template and passed to the Gemini model. The agent maintains a structured message history and a multi-turn conversation with the LLM to guide the reasoning process.

During inference, the agent interacts with our provided functions and tools via our specified \texttt{<execute>} tag~(e.g., \texttt{<execute>~dist(obj\_1,obj\_2)~</execute>}). These commands are parsed and then dispatched to a pre-defined set of spatial APIs, including distance estimation, object inclusion, relative positioning (left/right), and region queries (e.g., most left, middle). Execution results are returned to the LLM, which may iteratively refine its reasoning before producing the final answer, enclosed in \texttt{<answer>} tags when the LLM confirms the final answer.

The provided spatial APIs are a series of common functions and tools that can help the agent find out the answer. For simpler spatial relationships like left/right, we utilized the object mask centroid coordinate on the image to determine its spatial relationship. For more complex spatial relationships like distance estimation and determining whether an object is inside a specific region, we train a deep learning model to conduct end-to-end prediction. We introduce more details on these models in Sec.~\ref{subsec:distance} and Sec.~\ref{subsec:inside}.

\subsection{Distance Estimation Model}
\label{subsec:distance}
Since the AI City Challenge Physical AI Spatial Intelligence Warehouse dataset does not provide camera parameters or absolute depth information, we formulate the distance estimation task as a direct regression problem using only the available image and mask data. Given an RGB image \( I \in \mathbb{R}^{H \times W \times 3} \) and two binary object masks \( M_1, M_2 \in \{0,1\}^{H \times W} \), we aim to learn a model \( F \) that predicts the absolute distance \( D \in \mathbb{R} \) between the two objects. This is expressed as \( D = F(I, M_1, M_2) \). The model is trained to minimize the error between the predicted distance \( \hat{D} \) and the ground-truth distance \( D_{\text{gt}} \), using a standard regression L2 loss:
\[
\mathcal{L_{\text{dist}}} = \quad \left\| \hat{D} - D_{\text{gt}} \right\|_2^2
\]
We adopt ResNet-50~\cite{he2016deep} with 5 input channels that take an RGB image and two binary object masks as input.

In our experiments, we found that the model does not achieve satisfactory accuracy on smaller distance estimation~(less than 3m). To address this, we further fine-tuned another distance estimation model \(F_{small}\), which only trained on data with groundtruth distances smaller than 3m. In our final implementation, we cascade the two distance estimation models \(F\) and \(F_{small}\), whenever \(F\) predicts a value smaller than 3m, we use \(F_{small}\) to predict again, and use its prediction as our final answer for distance estimation questions.

\subsection{Inclusion Classification Model}
\label{subsec:inside}
In addition to distance estimation, we also design a binary classification model to determine whether one object is spatially included within another. Specifically, given an RGB image \( I \in \mathbb{R}^{H \times W \times 3} \) and two binary object masks \( M_1, M_2 \in \{0,1\}^{H \times W} \), we train a model \( G \) to predict an inclusion label \( y \in \{0, 1\} \), where \( y = 1 \) indicates that object A (corresponding to \( M_1 \)) is spatially included inside object B (corresponding to \( M_2 \)), and object B is required to be a buffer region object. The model is trained using the focal loss~\cite{lin2017focal} which can be expressed as:
\[
\mathcal{L}_{\text{inc}} = -\alpha_t (1 - p_t)^\gamma \log(p_t)
\]
where \( p_t \) is the model's predicted probability for the true class label \( y \), and \( \alpha_t \), \( \gamma \) are hyperparameters that control the focusing and weighting behavior of the loss. We follow the distance estimation model and use ResNet-50 as the backbone for the inclusion classification model.

%% file: figs/agent.tex
\begin{figure*}[t]
    \centering
    \includegraphics[width=0.98\linewidth]{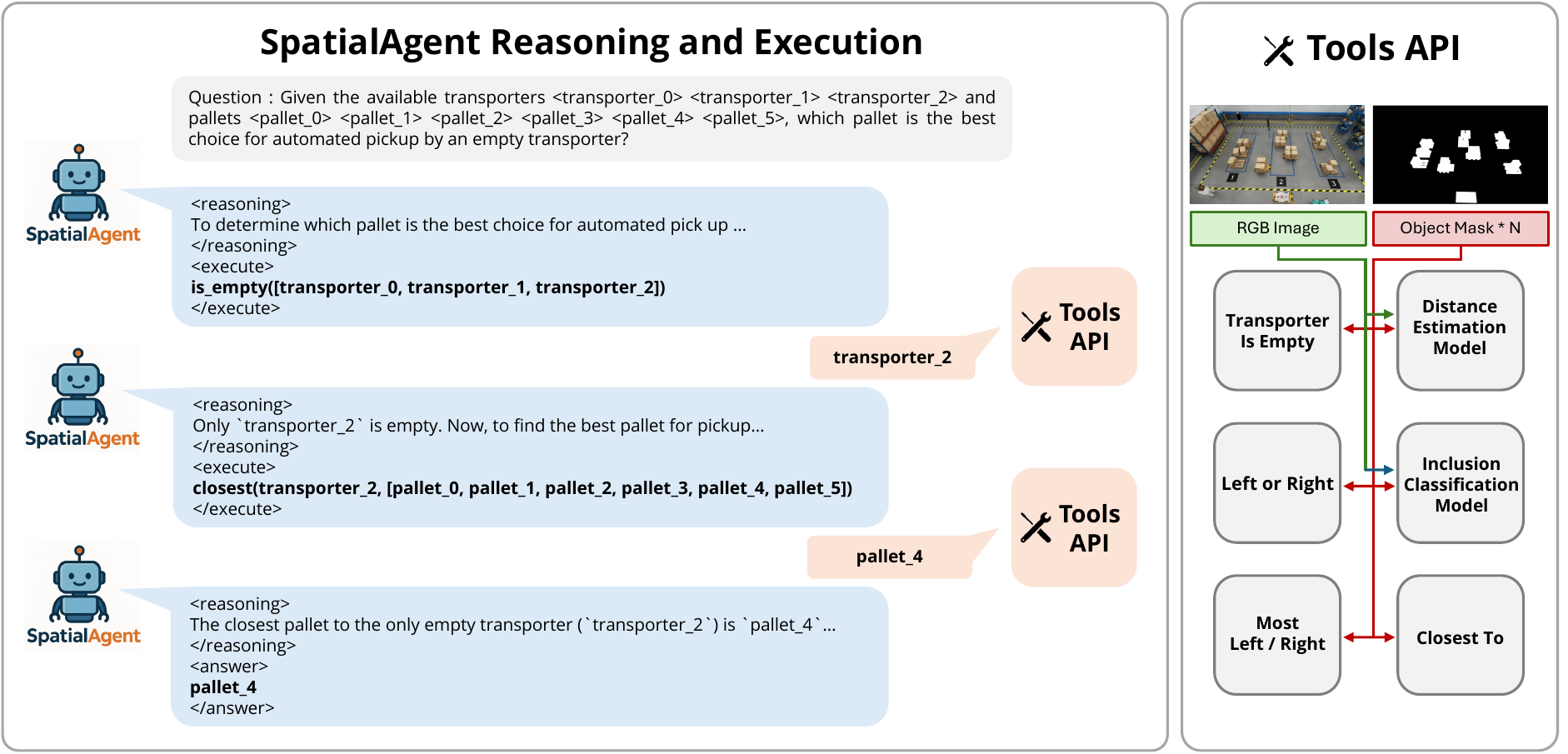}
    \caption{An illustration of our SpatialAgent framework.}
    \label{fig:agent}
\end{figure*}

%% file: sec/4_experiments.tex
\section{Experiments}
\label{sec:exp}

\subsection{Benchmark}
In our experiments, we utilize the 2025 AI City Challenge Physical AI Spatial Intelligence Warehouse dataset~\cite{Tang25AICity25}, a large-scale synthetic benchmark focusing on spatial reasoning and question answering within warehouse environments. This dataset is generated using the NVIDIA Omniverse platform and provides rich multimodal inputs, including RGB-D image pairs, object masks, and natural language QA pairs.

The QA pairs are categorized into four main types: spatial relations, multi-choice selection, distance estimation, and object counting. Each question is accompanied by both a free-form answer and a normalized single-word response for quantitative evaluation. The dataset comprises approximately 499K question and answer pairs for training, 1.9K for validation, and 19K for testing.

\subsection{Evaluation}
We follow the official evaluation protocol of the AI City Challenge Physical AI Spatial Intelligence Warehouse benchmark. The primary metric is the weighted average success rate across all question types. A prediction is considered successful if it satisfies the Acc@10 criterion~(within $\pm10\%$ of ground truth) for distance estimation and counting questions. For multi-choice and spatial relation questions, exact match accuracy is used.

\subsection{Implementation Details}

\paragraph{Question Pre-processing.}
To enable the LLM Agents understanding of the mask-to-object correspondence from the given question, we adopt a simple heuristic rule to pre-process the input question. We search the very first target object~(buffer, pallet, transporter, or shelf) that appears before each \texttt{<mask>}, and modify each \texttt{<mask>} to our specified object-aware format \texttt{<object\_ID>}~(e.g. \texttt{<buffer\_0>}, \texttt{<transporter\_1>}). After this pre-processing, the LLM can understand the mask-to-object relationship and conduct spatial reasoning and function calling. In some corner case questions where there is no preceding target object before \texttt{<mask>}, we query an LLM~\cite{comanici2025gemini} to rephrase the question to our specified object-aware format. 


\paragraph{Spatial Agent.}
We use Gemini-2.5-Flash~\cite{comanici2025gemini} as our LLM agent, building on top of Google Vertex API. We set the temperature to $0.2$ and other LLM parameters as default. During the question answering process, if the agent fails to successfully execute a function, we re-run by adding 128 tokens of thinking budget, this enables LLM to conduct more detailed reasoning and thus prevent format error.

\paragraph{Distance Estimation Model.}
We use ResNet-50 as our distance estimation model backbone, the training data is collected from the distance estimation question in the training set. A total of 245K data points collected from the training set are used to train our distance estimation model. We trained both distance estimation models for 5 epochs with L2 loss, using image resolution of (640, 480) and a learning rate of 1e-4. We also scale the groundtruth unit by 100~(from meter to centimeter), which helps to provide stronger supervision during training.

\paragraph{Inclusion Classification Model.}
We use ResNet-50 as our inclusion classification model, the training data is collected from the counting question's free-form answer in the training set. 158K data points collected from the training set are used to train our inclusion classification model. We trained the model for 5 epochs using focal loss, with an image resolution of (640, 480) and a learning rate of 1e-4.

\subsection{Performance}
We compared our performance with other teams on the 2025 AI City Challenge Track 3 leaderboard in Tab.~\ref{table:ranking}. Our proposed system achieve $95.86$\% accuracy on the testing set of the Physical AI Spatial Intelligence Warehouse Benchmark, ranking 1st place among all teams.
\input{table/ranking}

%% file: table/ranking.tex
\begin{table}[t]
\centering
\small
\begin{tabular}{clc}
\hline
Ranking & Team Name & Accuracy \\ \hline
\textbf{1} & \textbf{UWIPL\_ETRI (Ours)} & \textbf{95.8638} \\
2         & HCMUT.VNU     & 91.9735 \\
3         & Embia         & 90.6772 \\
4         & MIZSU         & 73.0606 \\
5         & HCMUS\_HTH    & 66.8861 \\
6         & MealsRetrieval& 53.4763 \\
7         & BKU22         & 50.3662 \\
8         & Smart Lab     & 31.9245 \\
9         & AICV          & 28.2993 \\
\hline
\end{tabular}
\caption{Leaderboard of the 9th AI City Challenge Track 3: Warehouse Spatial Intelligence.}
\label{table:ranking}
\end{table}

%% file: sec/6_conclusion.tex
\section{Conclusion}
\label{sec:conclusion}
In this work, we present a spatial understanding LLM agent system for complex indoor warehouse environments. Our system bridges the gap between perception and high-level reasoning by equipping a reasoning LLM with specialized API tools to support complex spatial question answering. Our method achieve state-of-the-art accuracy on the 2025 AI City Challenge Physical AI Spatial Intelligence Warehouse benchmark, demonstrating the effectiveness and generalizability of our approach.

\section{Acknowledgement}
This  work  was  supported  by  the  Electronics  and Telecommunications Research Institute (ETRI) grant funded by  the  Korean  Government  (Development  of  ICT Convergence Technology for Daegu-GyeongBuk Regional Industry) under Grant 25ZD1120. We also want to acknowledge and thank NCHC from Taiwan for providing the computing resources.